\documentclass[10pt,twocolumn,letterpaper]{article}

\usepackage{tcolorbox}

\usepackage[pagenumbers]{cvpr} 

\definecolor{cvprblue}{rgb}{0.21,0.49,0.74}
\usepackage[pagebackref,breaklinks,colorlinks,allcolors=cvprblue]{hyperref}

\def\paperID{} 
\def\confName{CVPR}
\def\confYear{2026}

\usepackage{soul} 
\usepackage{xcolor} 

\sethlcolor{yellow!20} 
\usepackage{tcolorbox}
\tcbset{colback=blue!10, colframe=blue!20, sharp corners} 

\usepackage{makecell}
\usepackage{multirow}
\usepackage{multicol}
\usepackage{stfloats} 
\title{From ``What'' to ``How'': \\ Constrained Reasoning for Autoregressive Image Generation}

\author{
Ruxue Yan$^{1}$ \quad Xubo Liu$^{1}$ \quad Wenya Guo$^{1}$ \quad Zhengkun Zhang$^{2}$ \quad Ying Zhang$^{1}$ \quad Xiaojie Yuan$^{1}$ \\
$^{1}$College of Computer Science, Nankai University \\
$^{2}$Baidu Inc. \\
\textbf{Correspondence:} yanruxue@dbis.nankai.edu.cn
}

\begin{document}
\maketitle
\begin{abstract}

Autoregressive image generation has seen recent improvements with the introduction of chain-of-thought and reinforcement learning.
%
%
However, current methods merely specify ``What'' details to depict by rewriting the input prompt, yet fundamentally fail to reason about ``How'' to structure the overall image.
%
This inherent limitation gives rise to persistent issues, such as spatial ambiguity directly causing unrealistic object overlaps.
%
%
To bridge this gap, we propose \textbf{CoR-Painter}, a novel framework that pioneers a ``\texttt{How-to-What}'' paradigm by introducing \underline{Co}nstrained \underline{R}easoning to guide the autoregressive generation.
%
%
Specifically, it first deduces ``\texttt{How} to draw'' by deriving a set of visual constraints from the input prompt, which explicitly govern spatial relationships, key attributes, and compositional rules. 
These constraints steer the subsequent generation of a detailed description (``\texttt{What} to draw''), providing a structurally sound and coherent basis for accurate visual synthesis.
%
%
%
%
%
Additionally, we introduce a \textbf{Dual-Objective GRPO} strategy that
specifically optimizes {the textual constrained reasoning and visual projection processes} to ensure the coherence and quality of the entire generation pipeline.
Extensive experiments on T2I-CompBench, GenEval, and WISE demonstrate that our method achieves state-of-the-art performance, with significant improvements in spatial metrics (e.g., +5.41\% on T2I-CompBench). 
%
%
%
%
%
\end{abstract}






\section{Introduction}
\label{sec:intro}

\begin{figure}[t]
  \centering
    \includegraphics[width=\linewidth]{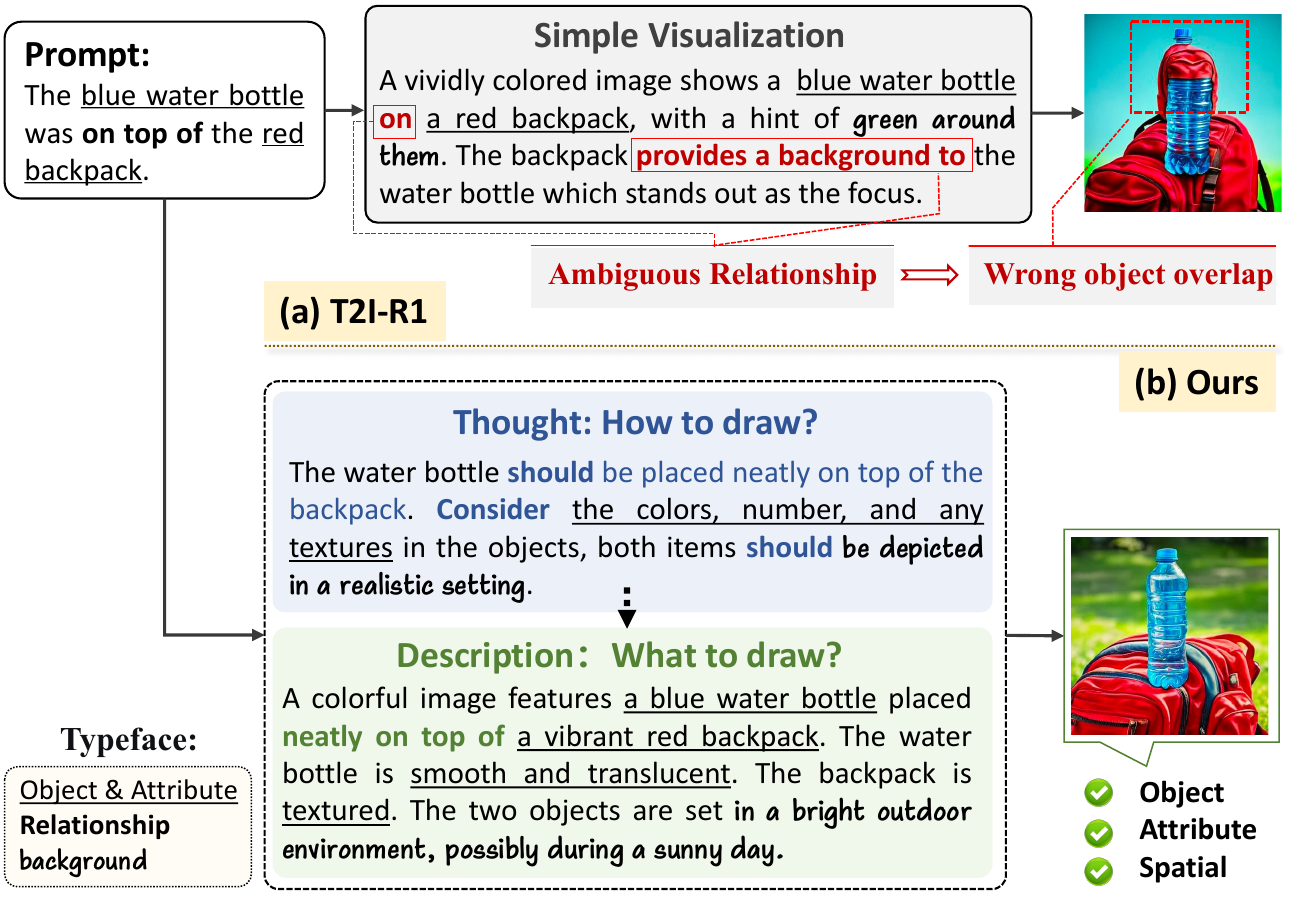}
  \caption{
  Illustration of the structure and output of (a) CoT-based method, T2I-R1 \cite{jiang2025t2i}, and (b) our CoR-Painter.
  }
  \label{intro}
\end{figure}

Text-to-image generation has emerged as a pivotal research area in computer vision.
%
%
The success of large language models (LLMs) \cite{DBLP:journals/corr/abs-2302-13971,DBLP:journals/corr/abs-2307-09288, DBLP:journals/corr/abs-2407-10671} has inspired autoregressive visual generation via next-token prediction. 
Concurrently, Chain-of-Thought (CoT) reasoning \cite{DBLP:conf/nips/Wei0SBIXCLZ22} has been extended to this domain, where methods first produce detailed descriptions of the input prompt and then optimize generated descriptions with reinforcement learning (RL) \cite{schulman2017proximal, rafailov2023direct, shao2024deepseekmath}, enhancing output quality and semantic consistency.



%
Existing CoT-based methods {\cite{wang2025mint, duan2025got, jiang2025t2i} primarily work by merely paraphrasing the input prompt into a more detailed description to guide image generation. 
However, this process simply focuses on the detailed expansion of ``What'' to depict, lacking the necessary reasoning about ``How'' the overall image should be coherently structured.
%
%
This inevitably introduces a critical flaw: it fails to establish a consistent generative blueprint, suffering from locally plausible details to conflict at the global level.
As demonstrated in Fig.~\ref{intro}(a), T2I-R1 \cite{jiang2025t2i} produces accurate details, such as the material, shape, and style of the water bottle and backpack, but it lacks a coherent set of spatial constraints to organize these elements into a logically consistent image. 
While individual relations like ``\textit{on}'' and ``\textit{provides a background to}'' are each reasonable in isolation, the absence of an overall structural constraint leads to the \textbf{ambiguous spatial interpretation} of the objects.
%
%
This ambiguity in the description directly propagates to the image synthesis process.
%
In an attempt to satisfy the ambiguous spatial cues simultaneously, the model is misled into generating multiple redundant objects that overlap in unreasonable arrangements, as highlighted in Fig.~\ref{intro}(a)—a clear \textbf{visual artifact}, severely compromising the semantic integrity of the image.

To address this issue, we propose \textbf{CoR-Painter}, a \underline{Co}nstrained \underline{R}easoning image generation framework, structured as a ``\texttt{How-to-What}'' paradigm, which explicitly introduces logically coherent and spatially well-organized constraints for the detailed descriptions.
Inspired by the human painting process, where an artist first establishes the composition and layout before adding details, 
CoR-Painter begins by clarifying ``\texttt{How}'' to organize the overall content and spatial structure of the image, forming a set of explicit constraints regarding the focus and arrangement. 
These constraints then guide the subsequent reasoning of ``\texttt{What}'' details to draw, leading to a well-structured and detailed visual description that serves as a direct mapping for image generation.
Driven by this ``\texttt{How-to-What}'' paradigm, CoR-Painter, as shown in Fig.~\ref{intro}(b), \textbf{first outputs constraint-based instructions} such as ``\textit{should be placed neatly on top of}'' to define \underline{spatial relations}, ``\textit{consider the colors, number, and textures}'' to highlight \underline{key visual aspects}, and ``\textit{should be depicted in an outdoor setting}'' to establish \underline{the background}. 
Guided by these constraints, the model \textbf{then generates a finely detailed description}: for example, ``\textit{a blue water bottle neatly on top of a vibrant red backpack}'', with further specifics on object appearance, such as a ``\textit{smooth}, \textit{translucent bottle}'' and a ``\textit{textured backpack}'', all set in a ``\textit{bright outdoor environment, possibly during a sunny day}''.
This structured constrained textual reasoning process ensures that the description is both globally consistent and locally detailed, effectively eliminating spatial ambiguity and object misplacement.
As a result, the generated image achieves higher semantic accuracy and visual coherence, as evidenced by the correct spatial arrangement and enhanced background depiction in Fig.~\ref{intro}(b).

In terms of model optimization, we adopt a \textbf{D}ual-\textbf{O}bjective GRPO (\textbf{DO-GRPO}) strategy to specifically optimize two core objectives:
%
%
%
%
1) The first objective is to strengthen semantic alignment and logical coherence in the textual reasoning process, ensuring that the model accurately understands and organizes the key information from the input prompt.
2) The second objective is to enhance visual–textual consistency in the image generation process, guiding the model to faithfully reproduce the detailed descriptions and spatial layouts during synthesis.
This dual-objective optimization effectively coordinates the reasoning and generation processes, leading to significant improvements in both semantic integrity and quality of final outputs.
%
%
Our contributions can be summarized as:


\begin{itemize}
    \item 
    We propose CoR-Painter, a ``\texttt{How-to-What}'' framework to ensure global coherence by prioritizing structural constraints (``\texttt{How}'') before visual detailing (``\texttt{What}'').
    \item 
    We introduce Dual-Objective GRPO to provide separate, dedicated rewards to reinforce the accuracy of the text reasoning and the fidelity of image generation.
    \item Extensive experiments conducted on T2I-CompBench, GenEval, and WISE demonstrate new state-of-the-art performance, especially in enhancing spatial relationships.
\end{itemize}
\section{Related Works}
\label{sec:related work}

\subsection{Autoregressive Models in Image Generation}
Inspired by the success of autoregressive generation in large language models \cite{DBLP:journals/corr/abs-2302-13971, DBLP:journals/corr/abs-2307-09288, DBLP:journals/corr/abs-2407-10671}, the autoregressive (AR) paradigm has gradually been applied to image generation, discretizing images into sequential tokens and generating them step by step through next-token prediction to produce high-quality images. 
LlamaGen \cite{sun2024autoregressive} is an early exploration, showing that autoregressive models can achieve competitive results in image generation.
%
Subsequent research has focused on improving efficiency, resolution, and semantic consistency.
%
AiM \cite{li2024scalable} leverages the Mamba architecture to optimize long-sequence modeling and accelerate inference; Token-Shuffle \cite{ma2025token} reduces token count to enable higher-resolution generation; CTF \cite{guo2025improving} applies coarse-to-fine token prediction to improve quality; GigaTok \cite{xiong2025gigatok} uses semantic regularization to handle complex visual tokenizers. Open-MAGVIT2 \cite{luo2024open} improves semantic consistency with super-large codebooks and sub-token prediction. More recent works propose next-scale/next-X \cite{ren2025beyond} prediction frameworks and AR models based on large-scale continuous tokens (NextStep-1) \cite{team2025nextstep}, achieving significant improvements in generation efficiency, quality, and high-resolution capability.
%
However, challenges remain in optimizing intermediate reasoning and ensuring semantic consistency in complex scenarios.

\begin{figure*}[t]
  \centering
    \includegraphics[width=0.95\linewidth]{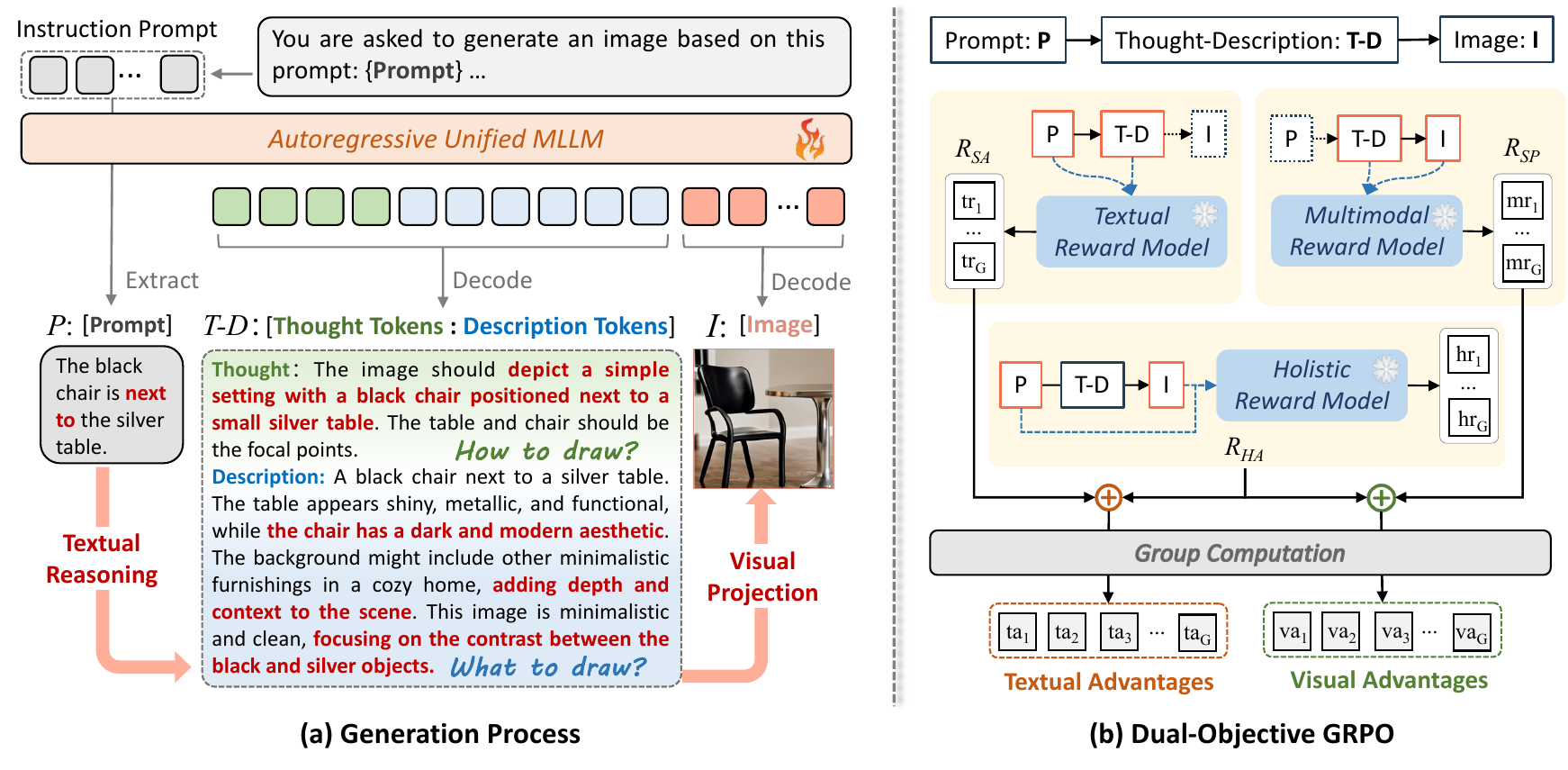}
  \caption{
  Overview of CoR-Painter: (a) illustration of the text-to-image generation process, and (b) Dual-Objective GRPO, 
  $R_\text{SA}$, $R_\text{SP}$ and $R_\text{HA}$ represent Semantic Anchoring Reward, Semantic Projection Reward and Holistic Alignment Reward, respectively.}
  \label{fig:pipeline}
\end{figure*}

\subsection{CoT Reasoning and RL in Image Generation}
%
Building on CoT reasoning in large language models, researchers have integrated reasoning and reinforcement learning (RL) into autoregressive image generation, advancing structured frameworks to guide the process.
%
%
BiCoT-GRPO \cite{jiang2025t2i} combines semantic- and token-level reasoning with generation rewards. PARM and PARM++ \cite{guo2025can} enhance autoregressive generation through stepwise evaluation using potential assessment rewards and reflection mechanisms. GoT \cite{fang2025got} and GoT-R1 \cite{duan2025got} integrate semantic-spatial reasoning and RL-based rewards to improve compositional and spatial alignment.
%
Research also focuses on reward mechanisms and training methods to reinforce reasoning-guided generation.
%
%
SUDER \cite{hong2025reinforcing} and CoRL \cite{jiang2025co} explore self-supervised dual rewards and co-reinforcement learning for multimodal optimization. FocusDiff \cite{pan2025focusdiff} uses RL for fine-grained text-image alignment, addressing semantic subtleties.
%
Despite these advancements, existing approaches rarely optimize textual reasoning and image generation separately. Our method addresses this by introducing objective-specific rewards for dedicated optimization while maintaining cross-modal consistency.

\section{Method}
In this section, we provide the details of CoR-Painter, starting with the generation process and then outlining how to train the model to achieve high-quality images with RL.

\subsection{Image Generation Pipeline}

As previously stated, our generation process follows the ``\texttt{How-to-What}'' paradigm for image generation.
Given an input prompt, we sequentially perform textual reasoning in terms of ``\texttt{How} to draw'' and ``\texttt{What} to draw'', producing constraint-guided instructions and structured logical descriptions that are then mapped to the image generation process, thereby serving as a bridge between linguistic understanding and the final visual rendering. 
The pipeline of this process is shown in Fig.~\ref{fig:pipeline}(a).
%

We use Janus-Pro \cite{chen2025janus} as the base model (\textit{i.e.}, the ``Auto-regressive Unified MLLM'') to jointly model text and image tokens within a shared representation space.
%
%
%
To guide the model in generation, 
we design an instruction prompt for the MLLM based on the original prompt. 
The designed prompt specifies the structure of the textual reasoning content by introducing specific tags that define two components: the \texttt{\textless thought\textgreater \textless /thought\textgreater} part and the \texttt{\textless description\textgreater \textless /description\textgreater} part. 
This design ensures that the generated reasoning aligns with the ``\texttt{How} to draw'' and ``\texttt{What} to draw'' processes.
The constructed instruction prompt is as follows:

\begin{tcolorbox}[colback=gray!5!white,colframe=gray!40!white,title=
Instruction Prompt for the Original Prompt 
, fonttitle=\bfseries, center title]

\small {You are asked to generate an image based on this prompt: ``\{\textit{Prompt}\}''. Let's think step by step.}

\small \texttt{<thought>}
\footnotesize \textsf{{\{List what objects, attributes, and spatial relations are mentioned in the prompt, noting any constraints they should follow.\}}}
\small\texttt{</thought>}\\
\small\texttt{<description>}
\footnotesize \textsf{\{Provide a brief, precise visualization of all elements in the prompt.\}}
\small \texttt{</description>}

\end{tcolorbox}

Following the above instruction prompt, the model first performs a global analysis based on the original prompt to determine the overall composition and semantic structure of the image, and then generates constraint-guided instructions as thought tokens for the primary objects, their attributes, and spatial relationships, specifying ``\texttt{How}'' they should be presented.
This explicit guidance provides prior global layout information for generating the description tokens regarding ``\texttt{What}'' to depict. It enables the model to progressively refine details in a structured manner and produce descriptive text that captures object appearances, relative positions, and scene atmosphere. 
This detailed and logical description is subsequently mapped to the image generation process.
%
This process of explicit thought followed by detailed description is referred to as \textbf{Textual Reasoning}, producing the sub-sequence $d$.
Next, an image generation start token \texttt{\textless img\_start\textgreater} is added, and the model generates the image $I$ based on $d$. This process is referred to as \textbf{Visual Projection}.
These two processes are subsequently optimized in a process-specific manner as follows.
\subsection{Dual-Objective GRPO}

\paragraph{Group Relative Policy Optimization (GRPO).}
GRPO is a reinforcement learning framework that enhances the reasoning capability of LMs by introducing group-based relative advantages instead of explicit value estimation.
Given a prompt $p$, the model samples $G$ responses $\{o_i\}_{i=1}^{G} \sim \pi_{\theta_{\text{old}}}(\cdot|p)$, each evaluated by a scalar reward $\mathcal{R}_i$. 
The advantage of each sample is normalized within the group:
\begin{equation}
A_i = \frac{\mathcal{R}_i - \text{mean}(\{\mathcal{R}_i\}_{i=1}^{G})}{\text{std}(\{\mathcal{R}_i\}_{i=1}^{G})}.
\end{equation}
The policy is then optimized through a clipped PPO-style objective with KL regularization:
\begin{equation}
\begin{split}
\mathcal{J}_{\text{GRPO}}(\theta) = \mathbb{E} \bigg[ & \frac{1}{G} \sum_{i=1}^{G} \min\left( r_i(\theta)A_i, \right. \\
& \left. \text{clip}(r_i(\theta),1-\varepsilon,1+\varepsilon)A_i \right) \\
& - \beta D_{\text{KL}}(\pi_{\theta} \parallel \pi_{\text{ref}}) \bigg],
\end{split}
\end{equation}
where $r_i(\theta) = \frac{\pi_{\theta}(o_i|p)}{\pi_{\theta_{\text{old}}}(o_i|p)}$ and $\beta$ controls the regularization strength.

\paragraph{Dual-Objective GRPO} 
While GRPO is originally designed for unimodal text reasoning, our task requires joint optimization over two heterogeneous modalities—textual reasoning and image generation.
To achieve this, we extend the GRPO framework into a two-objective reinforcement learning strategy, where each response $o_i$ consists of a textual reasoning sequence $s_i$ and an image sequence $t_i$, \textit{i.e.}, $o_i = (s_i, t_i)$.
The textual reasoning process corresponds to \textbf{Textual Reasoning}, producing the intermediate thought and description $d$, while the image generation process corresponds to \textbf{Visual Projection}, generating image tokens conditioned on $d$.
For each generation process, the token-wise probability ratio is defined as:
\begin{equation}
r_{i,j}(\theta) = 
\begin{cases} 
\dfrac{\pi_\theta(s_{i,j}|p,s_{i,<j})}{\pi_{\text{old}}(s_{i,j}|p,s_{i,<j})}, & 0 \leq j \leq |s_i|, \\[6pt]
\dfrac{\pi_\theta(t_{i,j}|p,s_i,t_{i,<j})}{\pi_{\text{old}}(t_{i,j}|p,s_i,t_{i,<j})}, & |s_i| < j \leq |s_i| + M.
\end{cases}
\end{equation}
%
This segmented ratio ensures that visual token generation is explicitly conditioned on the completed textual reasoning chain, maintaining semantic grounding across modalities.

\begin{figure*}[t]
  \centering
    \includegraphics[width=0.95\linewidth]{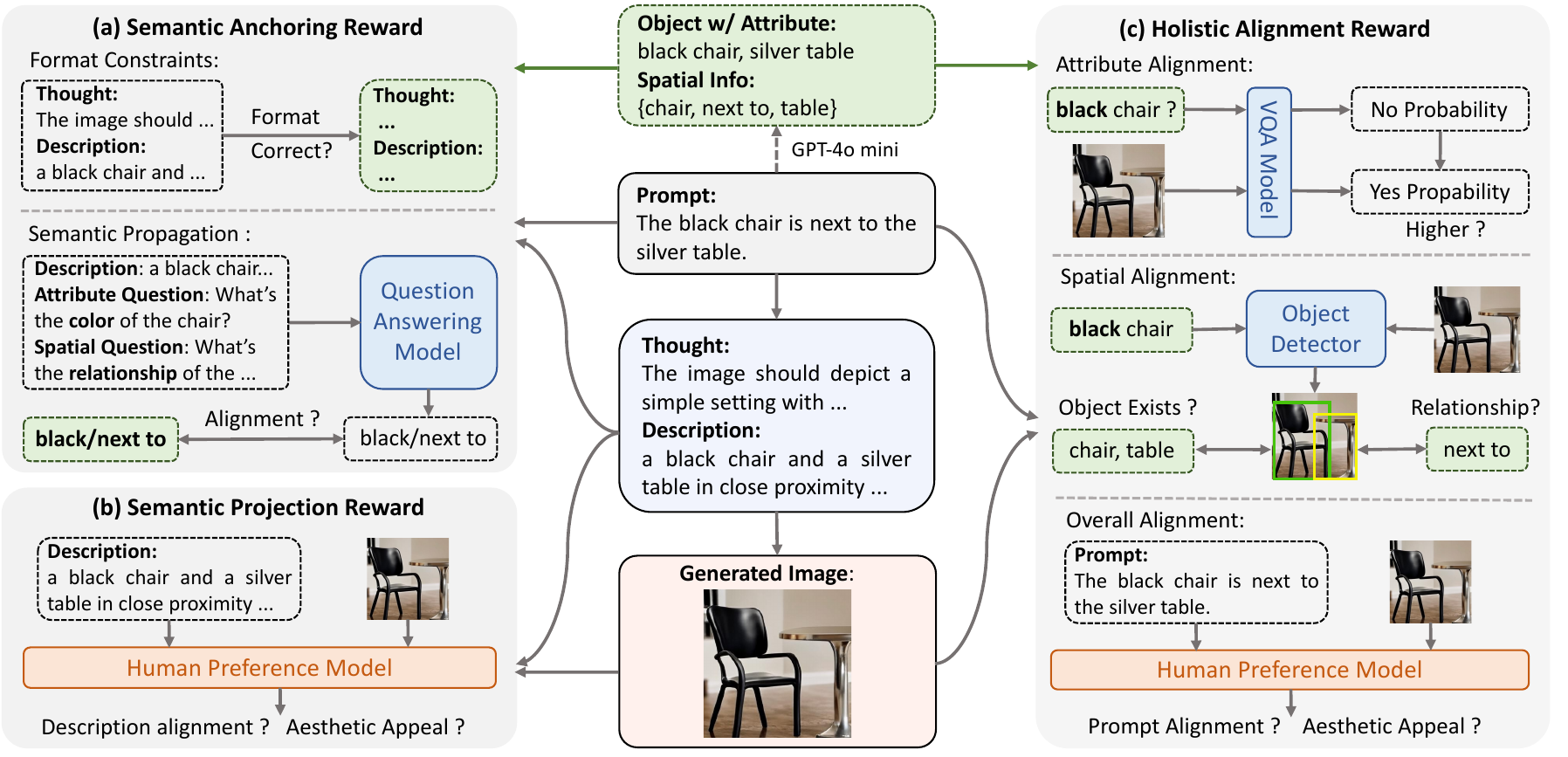}
  \caption{
   Overview of the reward computation.
  }
  \label{reward}
\end{figure*}

\paragraph{Objective-specific Reward Design.}
Based on above designs, Dual-Objective GRPO introduces modality-specific reward signals to guide each generation process, while maintaining shared advantage normalization for coherent optimization. 
As illustrated in Fig.\ref{fig:pipeline} (b), in the textual reasoning process, we apply the \textbf{\textit{Semantic Anchoring Reward}} ($R_{\text{SA}}$), computed by the textual reward model to measure the semantic consistency and logical faithfulness of the generated reasoning chain with respect to the original prompt. 
In the image generation process, the \textbf{\textit{Semantic Projection Reward}} ($R_{\text{SP}}$), provided by the multimodal reward model, evaluates how effectively the textual reasoning is projected into the visual domain. 
%
%
Additionally, the \textbf{\textit{Holistic Alignment Reward}} ($R_{\text{HA}}$) provides a global supervisory signal to capture the semantic alignment between the original prompt and the generated image. In practice, this reward is computed by three separate models that evaluate object attributes, spatial relationships, and overall scene semantics. To illustrate the reward as a whole, the pipeline depicts it with a single Holistic Reward Model, and the computations for each aspect are detailed in the following sections.
And the process-wise advantages are computed within each sampled group as:
\begin{equation}
\left\{
\begin{aligned}
A^{\text{text}}_i &= \frac{(R_{\text{SA}} + R_{\text{HA}}) - \text{mean}(\{R_{\text{SA}} + R_{\text{HA}}\}_{i=1}^{G})}{\text{std}(\{R_{\text{SA}} + R_{\text{HA}}\}_{i=1}^{G})}, \\[4pt]
A^{\text{img}}_i &= \frac{(R_{\text{SP}} + R_{\text{HA}}) - \text{mean}(\{R_{\text{SP}} + R_{\text{HA}}\}_{i=1}^{G})}{\text{std}(\{R_{\text{SP}} + R_{\text{HA}}\}_{i=1}^{G})},
\end{aligned}
\right.
\end{equation}
where $A^{\text{text}}_i$ and $A^{\text{img}}_i$ represent advantages for the textual reasoning and visual projection process, respectively.

\paragraph{Optimization Objective.}
The final objective of Dual-Objective GRPO is to optimize each process in a manner tailored to its respective advantages:

\begin{equation}
\begin{split}
&\mathcal{J}_{\text{Dual}}(\theta) 
= 
\\
&\mathbb{E} \Bigg[
\frac{1}{G} \sum_{i=1}^{G} 
\Big(\sum_{j=1}^{|s_i|}\!\min\!\left(r_{i,j}A^{\text{text}}_i, 
\text{clip}(r_{i,j},1-\varepsilon,1+\varepsilon)A^{\text{text}}_i\right) \\
&+\!\sum_{j=|s_i|+1}^{|s_i|+M}\!\min\!\left(r_{i,j}A^{\text{img}}_i, 
\text{clip}(r_{i,j},1-\varepsilon,1+\varepsilon)A^{\text{img}}_i\right)\!\Big) \\
&- \beta D_{\text{KL}}(\pi_{\theta} \parallel \pi_{\text{ref}})\!\Bigg].
\end{split}
\end{equation}
%
%
Through the dual-objective design, textual reasoning and visual projection are optimized according to their respective advantages, improving semantic coherence and visual fidelity while aligning reasoning and generation.

\subsection{Rewards Computation}
As mentioned above, our reward design has three components, each targeting a different aspect of generation. Key objects, attributes, spatial relations, and counts are pre-extracted from the prompt using GPT-4o-mini as deterministic references, enabling precise evaluation of how well the model preserves the prompt’s core semantics.

\begin{table*}[htbp]
  \centering
  \small
  \caption{Results on the T2I-CompBench and GenEval datasets.}
  \setlength{\tabcolsep}{3pt} 
    \begin{tabular}{lcccccc|ccccccc}
    \toprule
    \multirow{3}[6]{*}{Method} & \multicolumn{6}{c|}{T2I-CompBench}            & \multicolumn{7}{c}{GenEval} \\
 \cmidrule(lr){2-7} \cmidrule(lr){8-14}      & \multicolumn{3}{c}{Attribute Binding} & \multicolumn{2}{c}{Object Relation.} & \multirow{2}[4]{*}{Compl.} & \multirow{2}[4]{*}{{Overall}} & \multirow{2}[4]{*}{\makecell{Single \\ Obj.}} & \multirow{2}[4]{*}{\makecell{Two \\ Obj.}} & \multirow{2}[4]{*}{Count} & \multirow{2}[4]{*}{Color} & \multirow{2}[4]{*}{Pos.} & \multirow{2}[4]{*}{\makecell{Color \\ Attr.}} \\
\cmidrule(lr){2-4} \cmidrule(lr){5-6}         & Color & Shape & Texture & Spatial & N-Spatial &       &       &       &       &       &       &       &  \\
    \midrule
    \multicolumn{14}{c}{\textit{Diffusion-based Method }} \\
    \midrule
    SD-v1.5 \cite{rombach2022high}  & 37.58 & 37.13 & 41.86 & 16.27 & 31.12 & 32.16 & 0.43     & 0.97     & 0.38    & 0.35     & 0.76     & 0.04     & 0.06 \\
    SD-XL \cite{podell2023sdxl}  & 58.79 & 46.87 & 52.99 & 29.57 & 31.19 & 33.84 & 0.55     & 0.98    & 0.74    & 0.63     & 0.67    & 0.34    & 0.36 \\
    {PixArt-$\alpha$} \cite{chen2023pixart} & 66.90 & 49.27 & 64.77 & 20.64 & \textbf{31.97} & 34.33 & 0.48  & 0.98  & 0.50   & 0.44  & 0.80   & 0.08  & 0.07 \\
    SD3 \cite{DBLP:conf/icml/EsserKBEMSLLSBP24}  & 80.94 & 58.64 & 72.97 & 54.55 & 31.43 & 38.91 & -     & -     & -     & -     & -     & -     & - \\
    Stable v3 \cite{esser2024scaling} & 81.32 & 58.85 & 73.34 & 56.02 & 31.40 & 39.00  & -     & -     & -     & -     & -     & -     & - \\
    \midrule
    \multicolumn{14}{c}{\textit{AR-based Method}} \\
    \midrule
    LlamaGen \cite{sun2024autoregressive} & 42.02 & 39.67 & 51.03 & 19.75 & 30.50 & 30.81 & 0.32  & 0.71  & 0.34  & 0.21  & 0.58  & 0.07  & 0.04 \\
    Show-o \cite{DBLP:conf/iclr/XieMBZWLGCYS25} & 56.00  & 41.00  & 46.00  & 43.82 & 30.00   & 34.10 & 0.68  & 0.98  & 0.80   & \textbf{0.66}  & 0.84  & 0.31  & 0.50 \\
    Infinity \cite{DBLP:conf/cvpr/HanL0YZYPL25} & 73.79 & 46.50 & 59.19 & 40.67 & 30.76 & 38.06 & {0.73} & -     & 0.85  & -     & -     & 0.49  & 0.57 \\
    Janus-Pro-7B \cite{chen2025janus} & 63.59 & 35.28 & 49.36 & 35.96 & 30.85 & 37.42 & 0.80   & 0.99  & 0.89  & 0.59  & 0.90   & 0.79  & 0.66 \\
    DDT-LLaMA \cite{pan2025generative} & {72.80} & 51.40 & 64.20 & -     & -     & -     & 0.66  & 0.99  & 0.64  & 0.56  & 0.87  & 0.39  & 0.48 \\
    \midrule
    \multicolumn{14}{c}{\textit{AR-based Method + RL}} \\
    \midrule
    Show-o+PARM \cite{guo2025can} & 75.00  & 56.00  & 66.00  & 46.58 & 31.00  & 38.02 & 0.69  & 0.97  & 0.75  & 0.60   & 0.83  & 0.54  & 0.53 \\
    EMU3 \cite{wang2024emu3} & 75.44 & 57.06 & 71.64 & 52.33 & 30.84 & 37.00  & 0.64  & 0.99  & 0.76  & 0.38  & 0.85  & 0.45  & 0.40 \\
    GoT-R1-7B \cite{duan2025got} & 81.39 & 55.49 & 73.39 & 51.27 & 31.69 & 40.45 & 0.75  & 0.99  & 0.94  & 0.50   & 0.90   & 0.46  & 0.68 \\
    T2I-R1 \cite{jiang2025t2i} & 81.30 & 58.52 & 72.43 & 57.13 & 30.90 & 39.93 & 0.79  & 0.99  & 0.91  & 0.53  & 0.91  & 0.76  & 0.65 \\
    Janus-FocusDiff-7B \cite{pan2025focusdiff} & 83.00  & 60.30 & 72.80 & 58.43 & 30.99 & 40.98 & 0.85  & 0.99  & 0.95  & 0.63  & 0.93  & 0.85  & 0.75 \\
    CoR-Painter (Ours)  & \textbf{84.47} & \textbf{61.94} & \textbf{76.93} & \textbf{63.84} & 31.44 & \textbf{41.18} & \textbf{0.87} & \textbf{1.00} & \textbf{0.96} & 0.63  & \textbf{0.93} & \textbf{0.90} & \textbf{0.77} \\
    \bottomrule
    \end{tabular}%
  \label{tab:main-1}%
\end{table*}%

\paragraph{Semantic Anchoring Reward (SAR)} 
is used to ensure semantic fidelity in the textual reasoning process. 
As illustrated in Fig.~\ref{reward}(a), 
a format reward ensures the model performs ``\texttt{How} to draw'' reasoning by checking whether both thought and description tags are extracted, giving a higher score if both are present and lower if only one. The formula is:
%
\begin{equation}
R_{\text{Format}}(d_i) = w_1 \cdot f_{\text{thought}} + w_2 \cdot f_{\text{description}},
\end{equation}
%
where \( f_{\text{thought}} \) and \( f_{\text{description}} \) indicate whether valid tags are extracted, and the weights \( w_1 \), \( w_2 \) are both set to 0.5.

For semantic propagation, pre-extracted elements are turned into questions answered by the
QA model (Llama2-7b-chat \cite{touvron2023llama}), with reward measuring token-level consistency with reference answers:

\begin{equation}
R_{\text{Prop}}(d_i) = \frac{1}{K} \sum_{k=1}^{K} 
\text{match}(\text{QA}(d_i, q_k), a_k),
\end{equation}
where $q_k$ and $a_k$ denote the $k$-th question and reference answer, 
and $\text{match}(\cdot)$ computes a word-overlap or token-level similarity score.

The semantic anchoring reward $R_\text{SA}$ consists of the format reward and the semantic propagation reward:

\begin{equation}
R_{\text{SA}}(d_i) = R_{\text{Format}}(d_i) + R_{\text{Prop}}(d_i),
\end{equation}

\paragraph{Semantic Projection Reward (SPR)}  
is used to evaluate how faithfully the detailed description $d_i$ 
is realized in the generated image $I_i$, as well as the image’s aesthetic quality. 
As shown in Fig.\ref{reward}(b), we use a human preference model (HPSv2 \cite{wu2023human}) to compute both semantic alignment and aesthetic appeal:

\begin{equation}
R_{\text{SP}}(d_i, I_i) = \text{HPM}(d_i, I_i),
\end{equation}
where $\text{HPM}(\cdot)$ denotes the score given by HPSv2

\paragraph{Holistic Alignment Reward (HAR).}  
is used to evaluate the global cross-modal alignment between the prompt $p$ and generated image $I_i$, combining three assessments.
%
%
First, a VQA model (GIT) \cite{wang2022git} checks whether attribute questions derived from the prompt match the image content:
\begin{equation}
R_{\text{VQA}}(p, I_i) = \frac{1}{K} \sum_{k=1}^{K} P(\text{QA}(I_i, q_k) = a_k^{\text{ref}}),
\end{equation}
where $q_k$ and $a_k^{\text{ref}}$ are the $k$-th attribute question and reference answer, and $K$ is the total number of questions.
%
Second, GroundingDino \cite{liu2024grounding} detects prompt objects and checks their presence and spatial correctness in the image:
\begin{equation}
R_{\text{Det}}(p, I_i) = \frac{1}{N_o} \sum_{m=1}^{N_o} \text{match}(\text{Detect}(I_i, \text{obj}_m), \text{obj}_m^{\text{ref}}),
\end{equation}
where $q_k$ and $a_k^{\text{ref}}$ are the $k$-th attribute question and its reference, and $K$ is the total number of questions.
%
Finally, assessing the overall semantic and aesthetic alignment between the prompt and image:
\begin{equation}
R_{\text{Align}}(p, I_i) = \text{HPM}(p, I_i),
\end{equation}
where $\text{HPM}(\cdot)$ denotes the score given by the human preference model (HPSv2).
Formally, the ensemble reward is computed as:
\begin{equation}
R_{\text{HA}}(p, I_i) = R_{\text{VQA}} + R_{\text{Det}} + R_{\text{Align}}.
\end{equation}

\section{Experiments}

\subsection{Main Results}
We conduct experiments on 3 widely used benchmarks: 
T2I-Compbench \cite{DBLP:conf/nips/HuangSXLL23}, GenEval \cite{DBLP:conf/nips/GhoshHS23} and WISE \cite{niu2025wise}.
The results are listed as follows:

\noindent \textbf{For the T2I-Compbench dataset} \cite{DBLP:conf/nips/HuangSXLL23},
%
we evaluate generated images on attributes, object relationships, and complex scenes. Replacing the original UniDet \cite{zhou2022simple} with the open-vocabulary GroundingDINO \cite{liu2024grounding} substantially improves results—for example, T2I-R1’s spatial score rises from 33.78\% to 57.13\%, demonstrating the detector’s effectiveness.
As shown in Tab.~\ref{tab:main-1}, our CoR-Painter outperforms existing methods across nearly all categories, with a notable 5.41\% gain on the spatial relationship metric, demonstrating strong capabilities of layout reasoning and accurate object positioning. The non-spatial score is slightly lower than PixArt-$\alpha$ ($\downarrow$0.0025), which is negligible given similar CLIP scores (0.30–0.31). Overall, these results fully demonstrate the effectiveness of our generative reasoning chain combined with DO-GRPO.


\begin{table}[tbp]
  \centering
  \small
  \caption{Results on the WISE dataset.}
  \setlength{\tabcolsep}{2.2pt} 
    \begin{tabular}{lccccccc}
    \toprule
    Method & Cult. & Time  & Space & Biol. & Phys. & Chemi. & All \\
    \midrule
    VILA-U \cite{wu2024vila} & 0.26  & 0.33   & 0.37  & 0.35   & 0.39  & 0.23   & 0.31 \\
    Show-o \cite{DBLP:conf/iclr/XieMBZWLGCYS25} & 0.28  & 0.4   & 0.48  & 0.30   & 0.46  & 0.30  & 0.35 \\
    EMU3 \cite{wang2024emu3} & 0.34  & 0.45  & 0.48  & 0.41  & 0.45  & 0.27  & 0.39 \\
    Janus-Pro-7B \cite{chen2025janus} & 0.30   & 0.37  & 0.49  & 0.36  & 0.42  & 0.26  & 0.35 \\
    T2I-R1 \cite{jiang2025t2i} & 0.56  & 0.55  & 0.63  & 0.54  & 0.55  & 0.30   & 0.54 \\
    CoR-Painter (Ours)  & \textbf{0.61} & \textbf{0.56}  & \textbf{0.66}  & \textbf{0.55}  & \textbf{0.64}  & \textbf{0.33}  & \textbf{0.58} \\
    \bottomrule
    \end{tabular}%
  \label{tab:wise}%
\end{table}%

\begin{table*}[htbp]
  \centering
  \small
  \caption{Ablations on reasoning chain structure and reward design.}
    \begin{tabular}{lllcccccc}
    \toprule
    \multirow{2}[4]{*}{Components} &       &       & \multicolumn{3}{c}{Attribute Binding} & \multicolumn{2}{c}{Object Relationship} & \multirow{2}[4]{*}{Complex} \\
\cmidrule{4-8}          &       &       & Color & Shape & Texture & Spatial & Non-Spatial &  \\
    \midrule
    \multicolumn{3}{l}{Janus-Pro-7B (baseline) \cite{chen2025janus}} & 63.59 & 35.28 & 49.36 & 35.96 & 30.85 & 37.42 \\
    \midrule
    \multicolumn{3}{l}{\textbf{Ours-Full Chain}} & \textbf{84.47} & \textbf{61.94} & \textbf{76.93} & \textbf{63.84} & \textbf{31.44} & \textbf{41.18} \\
    \multicolumn{3}{l}{\quad (a) w/o thought (``\texttt{How} to draw'')} & 81.14 & 58.04 & 74.92 & 60.85 & 30.90 & 40.32 \\
    \multicolumn{3}{l}{\quad (b) w/o description (``\texttt{What} to draw'')} & 78.99 & 53.00 & 69.03 & 54.66 & 30.53 & 39.02 \\
    \midrule
    \midrule
    \textbf{Ours-Full Rewards} & & & \textbf{84.47} & \textbf{61.94} & \textbf{76.93} & \textbf{63.84} & \textbf{31.44} & \textbf{41.18} \\
    \quad (c) w/o SAR: HAR$\Leftrightarrow$SPR+HAR &   & & 83.46 & 61.17 & 75.23 & 62.89 & 30.57 & 40.94 \\
    \quad (d) w/o SPR: SAR+HAR$\Leftrightarrow$HAR &  &    & 82.25 & 60.58 & 74.85 & 61.25 & 29.85 & 40.08 \\
    \quad (e) w/o SAR\&SPR: HAR$\Leftrightarrow$HAR &   &   & 81.31 & 59.38 & 73.85 & 59.89 & 29.27 & 39.51 \\
    \midrule
    \midrule
    \multicolumn{3}{l}{\textbf{Ours-Full HAR}} & \textbf{84.47} & \textbf{61.94} & \textbf{76.93} & \textbf{63.84} & \textbf{31.44} & \textbf{41.18}  \\
    \multicolumn{3}{l}{\quad (f) w/o VQA} & 81.95 & 58.65 & 73.42 & 63.27 & 30.92 & 39.96 \\
    \multicolumn{3}{l}{\quad (g) w/o Det} & 82.00 & 60.25 & 72.90 & 58.38 & 30.99 & 39.47 \\
    \multicolumn{3}{l}{\quad (h) w/o HPM} & 81.74 & 59.38 & 73.85 & 60.89 & 29.79 & 40.02 \\
    \bottomrule
    \end{tabular}%
  \label{tab:ablation}%
\end{table*}%

\noindent \textbf{For the GenEval dataset}, we evaluate our model across a diverse set of tasks, including single-object, two-object, counting, color, position, and attribute-binding scenarios. As presented in Fig.~\ref{tab:main-1}, CoR-Painter achieves SOTA in overall performance, and notably surpasses the previous SOTA method, Janus-FocusDiff, by 5\% on the spatial positioning task. 
Our approach emphasizes global layout coherence over strict quantity control, leading to some fluctuation in the counting metric, which is slightly lower than show-o. Overall, CoR-Painter performs well on this benchmark, accurately organizing object positions while maintaining semantic and visual consistency.

\noindent \textbf{On the WISE dataset} \cite{niu2025wise}, text prompts require domain knowledge and common-sense reasoning, including time, geography, and natural science, to generate accurate images. This places higher demands on the model’s reasoning ability. Previous models focus on rewriting prompts into detailed descriptions, but for implicit prompts requiring knowledge reasoning, they fail to infer the objects that need to be expanded, resulting in poor performance. In contrast, CoR-Painter infers the target objects through a constraint-guided process, enabling correct image generation. As shown in Tab.~\ref{tab:wise}, CoR-Painter outperforms existing methods, demonstrating its effectiveness in knowledge-driven image generation.

\begin{figure}[t]
  \centering
    \includegraphics[width=\linewidth]{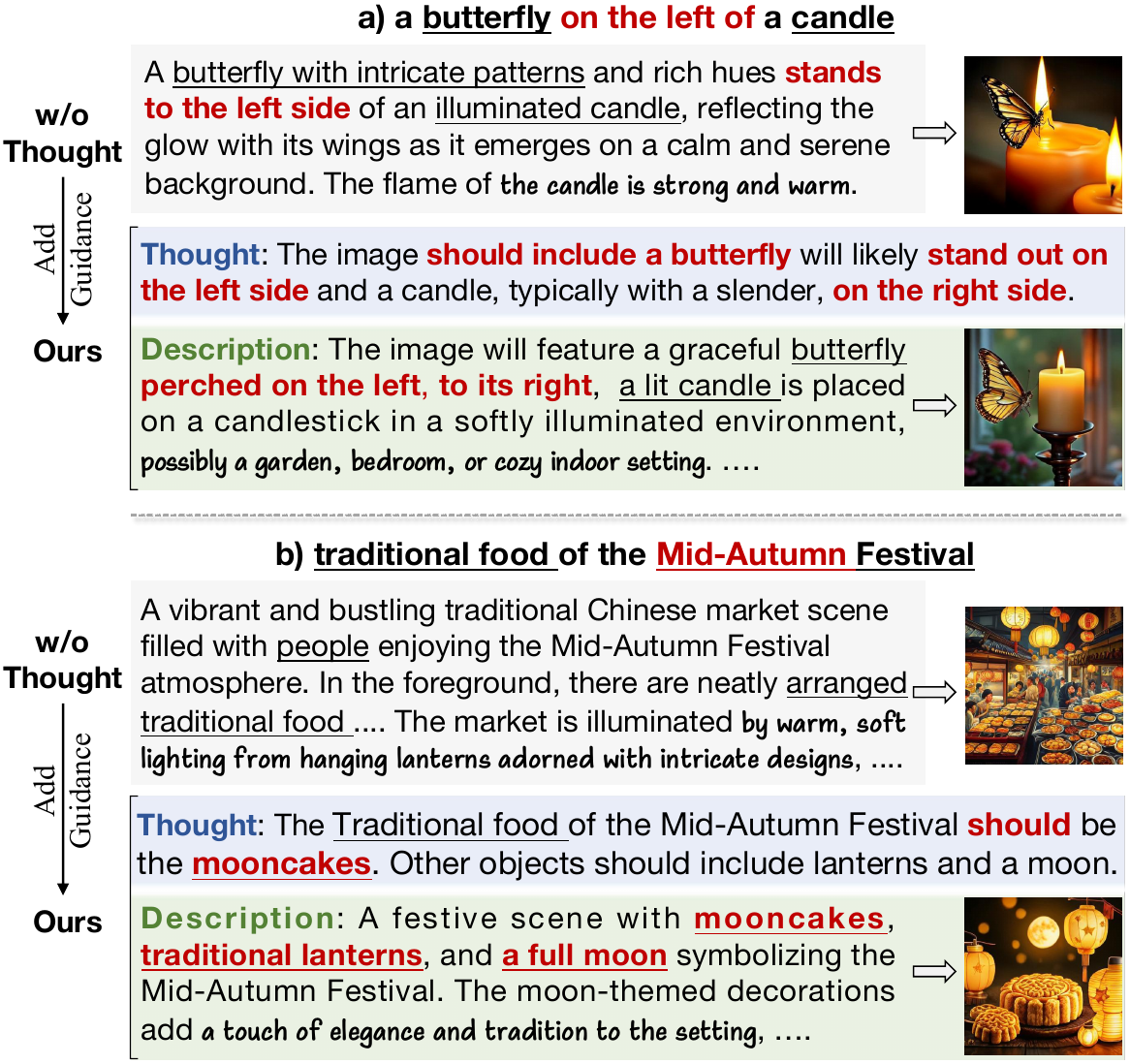}
  \caption{
  Comparison of description and generated images before and after incorporating the thought process.
  }
  \label{fig:ablation-thought}
\end{figure}

\begin{figure*}[t]
  \centering
    \includegraphics[width=0.95\linewidth]{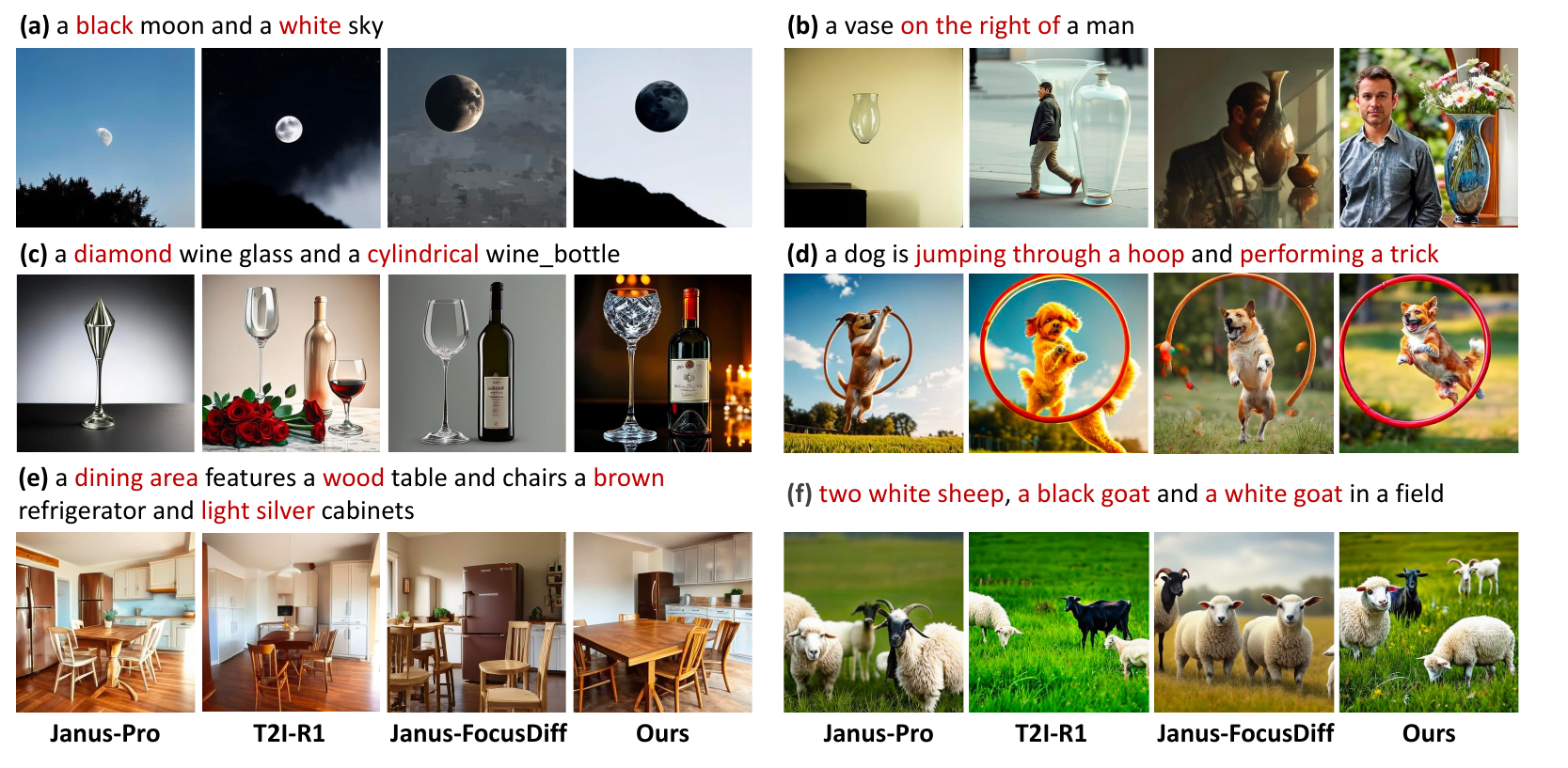}
  \caption{
  Qualitative comparison of CoR-Painter against recent methods on typical cases.
  }
  \label{fig:case}
\end{figure*}

\begin{figure}[h]
  \centering
    \includegraphics[width=0.9\linewidth]{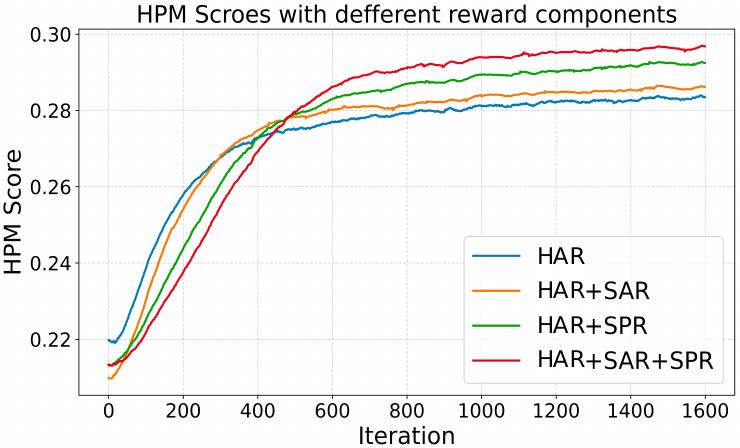}
  \caption{
  HPM scores 
  under different reward ablations.
  }
  \label{fig:ablation-rewards}
\end{figure}

\subsection{Ablation Study}
To further verify the effectiveness of the ``\texttt{How-to-What}'' paradigm and DO-GRPO, we perform ablations on the reasoning chain structure and reward design.

\noindent \textbf{Reasoning chain structure}. 
As shown in Tab.~\ref{tab:ablation} (rows a–b), removing the thought process reduces semantic accuracy, while omitting the description process degrades performance due to a lack of structured mappings. Both components are essential for semantic fidelity and spatial consistency. Visualization in Fig.~\ref{fig:ablation-thought} shows that the thought process clarifies object properties and relationships, \textit{e.g.}, the butterfly and candle—and improves reasoning for implicit objects, like ``\textit{Mid-Autumn Festival traditional food should be mooncakes}'', producing more complete and consistent output images.

\noindent \textbf{Reward design}. 
As shown in Tab.~\ref{tab:ablation} (c)–(e), removing SAR weakens semantic transfer, affecting both attribute and spatial performance, while removing SPR reduces text–image alignment. When both are removed, leaving only HAR, performance drops significantly. Ablations on individual HAR rewards (rows (f)–(h)) also show clear declines, further highlighting the importance of each reward. Qualitative analysis of the HPM reward reveals that it reflects prompt–image alignment and aesthetics. As shown in Fig.~\ref{fig:ablation-rewards}, HPM scores are highest with both SAR and SPR, and drop when any reward is removed. These results show that the dual-objective design effectively optimizes text and image generation, improving reasoning, image quality, and spatial relationships.

\subsection{Case Study}
We analyze cases to intuitively demonstrate the effectiveness of our method compared with three recent approaches: Janus-Pro (our baseline), T2I-R1 (previous SOTA CoT-based method), and Janus-FocusDiff (previous SOTA AR-based method). 
%
%
As shown in Fig.~\ref{fig:case}, Janus-Pro fails to generate some target objects, resulting in incomplete images, such as only producing the vase in case (b).
T2I-R1 produces rich visual details but suffers from unreasonable layouts and incorrect object counts, \textit{e.g.}, the spatial relation between the person and vase in case (b), and generating two wine bottles instead of one in case (c).
Janus-Focus handles basic details well but struggles in complex scenes, \textit{e.g.}, incorrect cabinet colors in case (e) and errors with object quantities in case (f).
%
%
%
In contrast, our CoR-Painter first generates constraints for global elements and their positions to guide the subsequent detailed descriptions. This enables precise attribute generation, such as the color and texture of objects in cases (a) and (c), and coherent spatial layouts, such as the positions of the person and vase in case (b). It also performs well in complex scenes with multiple objects, as in case (e), and accurately handles varying numbers and colors of objects, such as the goats and sheep in case (f).

\section{Conclusion}

In this paper, we propose CoR-Painter, a novel framework that pioneers a
``\texttt{How-to-What}'' paradigm to enhance text-to-image generation. 
By first generating directive constraints on ``\texttt{How} to draw'' and then reasoning about ``\texttt{What} to draw'', our approach produces structured and detailed visual descriptions, effectively resolving unclear spatial relationships and misaligned object placements.
%
%
We further introduce DO-GRPO to optimize text and image generation separately, improving overall quality and semantic consistency. 
Extensive experiments on multiple benchmarks show that our method consistently outperforms existing approaches, demonstrating its effectiveness in producing high-quality, semantically aligned images.  

\clearpage
{
    \small
    \bibliographystyle{ieeenat_fullname}
    \bibliography{main}
}
\clearpage
\setcounter{page}{1}
\maketitlesupplementary


\section{More Experiment Details}

\subsection{Benchmark}
We conduct systematic evaluation of our method using the following three authoritative benchmarks: T2I-Compbench \cite{DBLP:conf/nips/HuangSXLL23}, GenEval \cite{DBLP:conf/nips/GhoshHS23} and WISE \cite{niu2025wise}. These benchmarks allow us to thoroughly test key capabilities such as compositional text understanding and world knowledge reasoning.

\paragraph{T2I-CompBench} employs 6,000 carefully designed text prompts to specifically evaluate model compositional reasoning capabilities. The benchmark's testing framework covers three core dimensions: attribute binding, object relationships, and complex compositions. Each dimension is further subdivided into six subcategories, examining model performance in color, shape, texture, spatial relationships, non-spatial relationships, and complex composition tasks respectively.

\paragraph{GenEval} employs 553 systematically designed prompts across 6 core tasks to evaluate fine-grained text-to-image alignment. This object-focused benchmark assesses capabilities in single/multiple object generation, counting, color binding, spatial relationships, and attribute binding, using templates populated with MS-COCO objects and basic attributes.

\paragraph{WISE} focuses on evaluating model capabilities in world knowledge integration and reasoning. This benchmark contains 1,000 test prompts covering domains such as cultural common sense, spatiotemporal reasoning, and natural sciences. Strong performance on this benchmark requires models to not only understand literal meanings but also infer the specific real-world scenarios or entities described in the text. To obtain more consistent evaluation results, we made appropriate adjustments to the original evaluation instructions.

\subsection{Prompt settings}
To more clearly illustrate our generation pipeline, we provide the full instruction prompt used during both training and inference. The prompt guides the model through the two processes of ``\texttt{How to draw}'' and ``\texttt{What to draw}'', establishing a structured reasoning path for image synthesis. The complete prompt used in our experiments is presented here:

\begin{tcolorbox}[colback=gray!5!white,colframe=gray!40!white,title=
Instruction Prompt for the Original Prompt 
, fonttitle=\bfseries, center title]

\small {You are asked to generate an image based on this prompt: ``\{\textit{Prompt}\}''. Let's think step by step about what should appear in the image.}

\small \texttt{<thought>}
\footnotesize \textsf{{\{List what objects, attributes, and spatial relations are mentioned in the prompt, noting any constraints they should follow. Keep it concise, factual, and focused on concrete visual details — around 2–3 short sentences.\}}}
\small\texttt{</thought>}\\
\small\texttt{<description>}
\footnotesize \textsf{\{Provide a brief, precise visualization of all elements in the prompt. Include every object, specify visual attributes (color, number, shape, texture) if given, and clarify relationships (e.g., spatial) between them. Be concise ($\leq 50$ words) and do not add anything beyond the prompt.\}}
\small \texttt{</description>}

\small
Output explicitly as:\\
Thought: [content of thought]\\
Description: [content of description]

\end{tcolorbox}

\subsection{Training Settings.} 

Our experimental configuration employs a rigorously curated dataset comprising 6,786 text prompts exclusively derived from the training partitions of T2I-CompBench and the methodology established in \cite{guo2025can}. The complete training corpus consists of precisely prompts, with no visual data incorporated throughout the training procedure. 
\begin{table}[h]
\centering
\caption{Model Architecture and Optimization Parameters}
\begin{tabular}{lc}
\hline
\textbf{Parameter} & \textbf{Value} \\
\hline
Learning Rate & 1e-6 \\
Beta Coefficient $\beta$ & 0.01 \\
Group Size $G$ & 8 \\
Classifier-Free Guidance Scale & 5 \\
Max Gradient Norm & 1.0 \\
Batch Size & 8 \\
Training Steps & 1,600 \\
Gradient Accumulation Steps & 2 \\
Image Resolution $h \times w$ & $384 \times 384$ \\
\hline
\end{tabular}
\label{tab:model_architecture_params}
\end{table}

\begin{figure*}[h]
  \centering
    \includegraphics[width=0.85\linewidth]{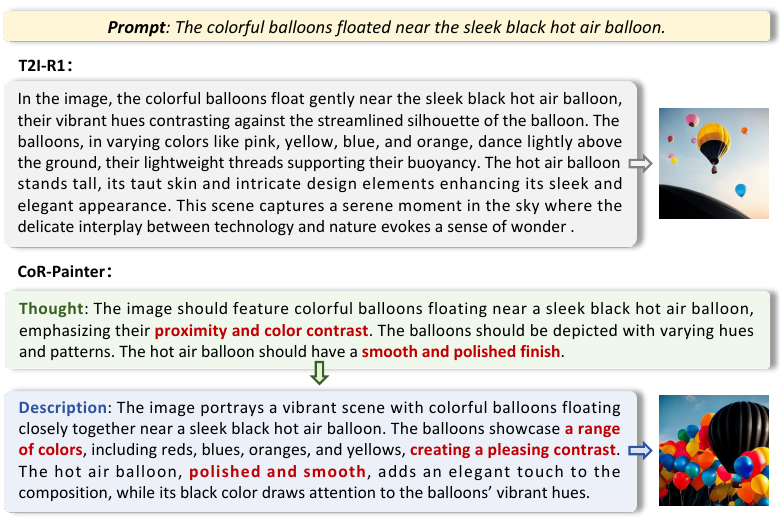}
  \caption{ Detailed comparisons between T2I-R1 and our CoR-Painter.
 %
  }
  \label{fig:case1}
\end{figure*}

\section{More Qualitative Results}
\subsection{Detailed Intuitive Comparisons}
We provide additional detailed visual comparisons between our CoR-Painter and the CoT-based SOTA method T2I-R1 to intuitively demonstrate the differences and advantages in image generation.
These comparisons cover diverse prompt types, including spatial relationships, multi-object attribute binding, knowledge-based reasoning, and complex scene composition, highlighting the model’s capability in handling various prompts.
As shown in Fig.~\ref{fig:case1}-\ref{fig:case4}, the direct comparison between textual reasoning processes and their corresponding image illustrates how CoR-Painter systematically guides description generation through constraint instructions produced during the thought process, ensuring proper organization of object attributes, counts, and spatial layout.
This ``\texttt{How-to-What}'' paradigm not only provides explicit guidance on target objects and spatial arrangements but also maintains logical coherence in knowledge-intensive scenarios. The resulting images show clear improvements in semantic accuracy, spatial relationships, and visual consistency, effectively mitigating typical issues such as object overlap and counting errors caused by ambiguous layouts, while enhancing overall scene completeness and interpretability.

\subsection{More Thought Ablation Visualization}
As shown in Fig.~\ref{fig:case5}-\ref{fig:case8}, we compare the textual reasoning processes and corresponding generation images without and with the thought process. 
To clearly illustrate the input for the ablation study, we also provide the full prompt used in the version without the thought process, allowing readers to directly see what information the model receives and how it affects the generated descriptions and images.

\begin{tcolorbox}[colback=gray!5!white,colframe=gray!40!white,title=
Ablation Instruction Prompt
, fonttitle=\bfseries, center title]

\small {You are asked to generate an image based on this prompt: ``\{\textit{Prompt}\}''. Let's think step by step about what should appear in the image.}

\small\texttt{<description>}
\footnotesize \textsf{\{Provide a brief, precise visualization of all elements in the prompt. Include every object, specify visual attributes (color, number, shape, texture) if given, and clarify relationships (e.g., spatial) between them. Be concise ($\leq 50$ words) and do not add anything beyond the prompt.\}}
\small \texttt{</description>}

\small
Output explicitly as:\\
Description: [content of description]

\end{tcolorbox}
These visual results elucidate how the constraint instructions generated during the thought process ensure coherent spatial layouts and consistent logical relationships among objects, clearly highlighting the contribution of thought-guided reasoning to overall image quality.
When this module is removed, the resulting descriptions lack structural guidance, leading to significantly degraded spatial arrangements and disrupted inter-object relationships. This ablation study visually substantiates the critical role of our ``\texttt{How-to-What}''  paradigm in maintaining structural integrity and semantic consistency in the generated images.

\begin{figure*}[t]
  \centering
    \includegraphics[width=0.85\linewidth]{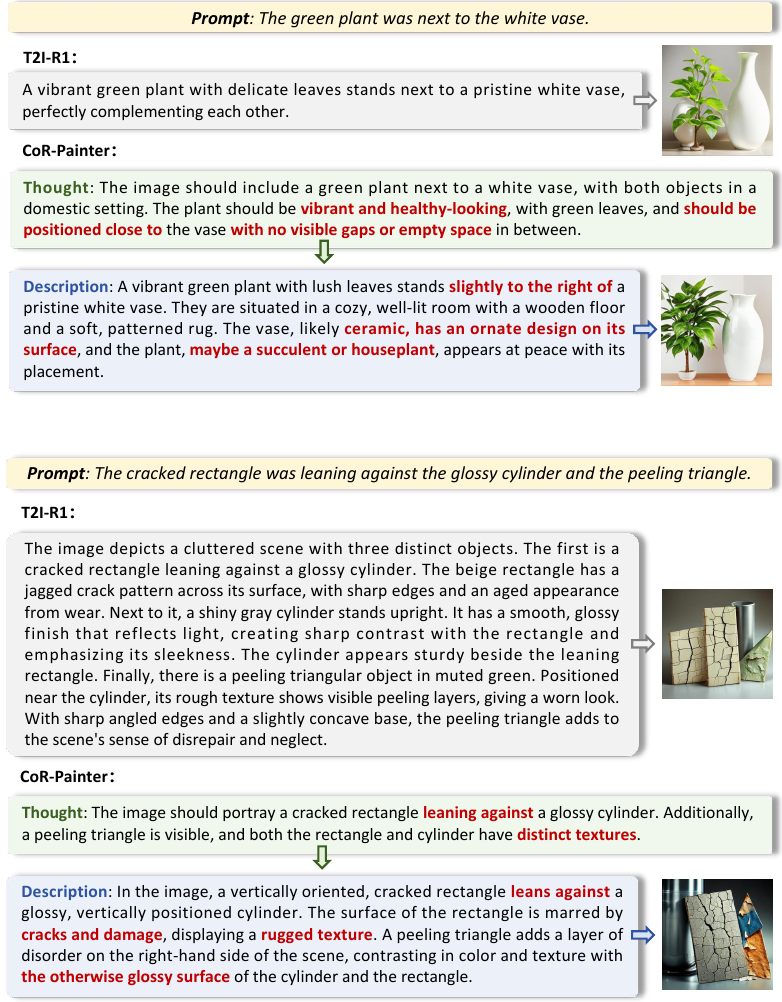}
  \caption{  Detailed comparisons between T2I-R1 and our CoR-Painter.
 %
   }
  \label{fig:case2}
\end{figure*}

\begin{figure*}[t]
  \centering
    \includegraphics[width=0.85\linewidth]{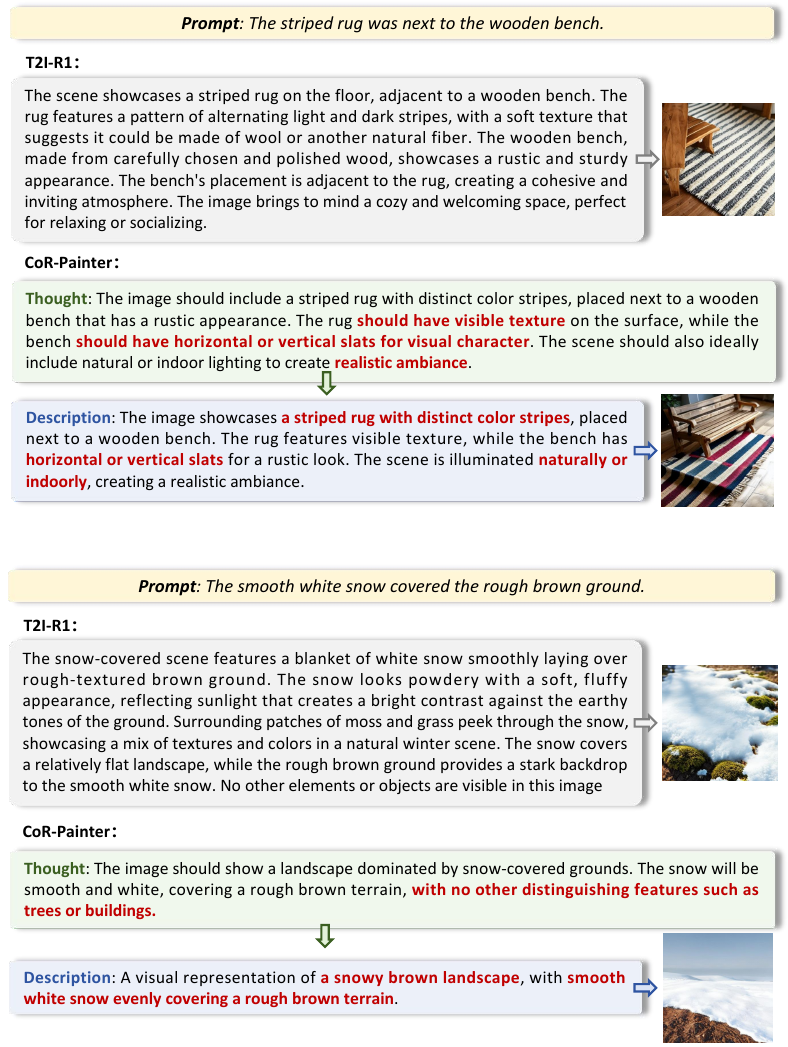}
  \caption{  Detailed comparisons between T2I-R1 and our CoR-Painter.
 %
  }
  \label{fig:case3}
\end{figure*}

\begin{figure*}[t]
  \centering
    \includegraphics[width=0.85\linewidth]{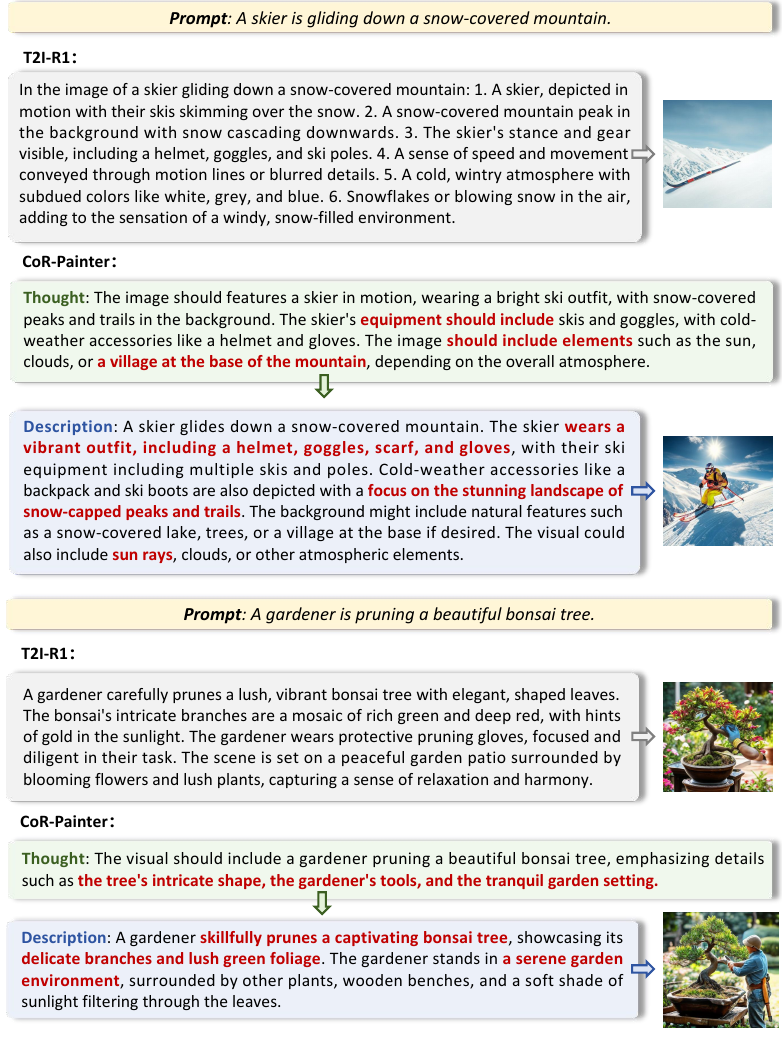}
  \caption{  Detailed comparisons between T2I-R1 and our CoR-Painter.
 %
  }
  \label{fig:case4}
\end{figure*}

\begin{figure*}[t]
  \centering
    \includegraphics[width=0.85\linewidth]{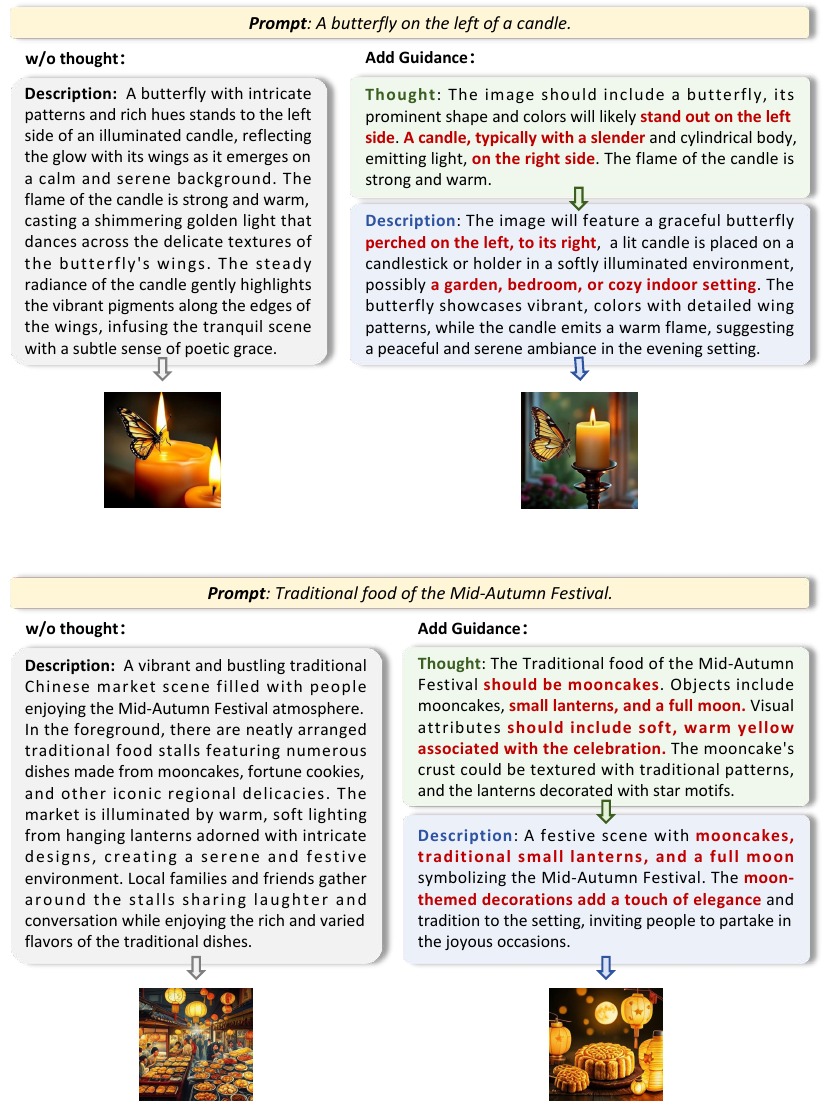}
  \caption{ Comparisons of images with and without the thought process.
 %
  }
  \label{fig:case5}
\end{figure*}

\begin{figure*}[t]
  \centering
    \includegraphics[width=0.85\linewidth]{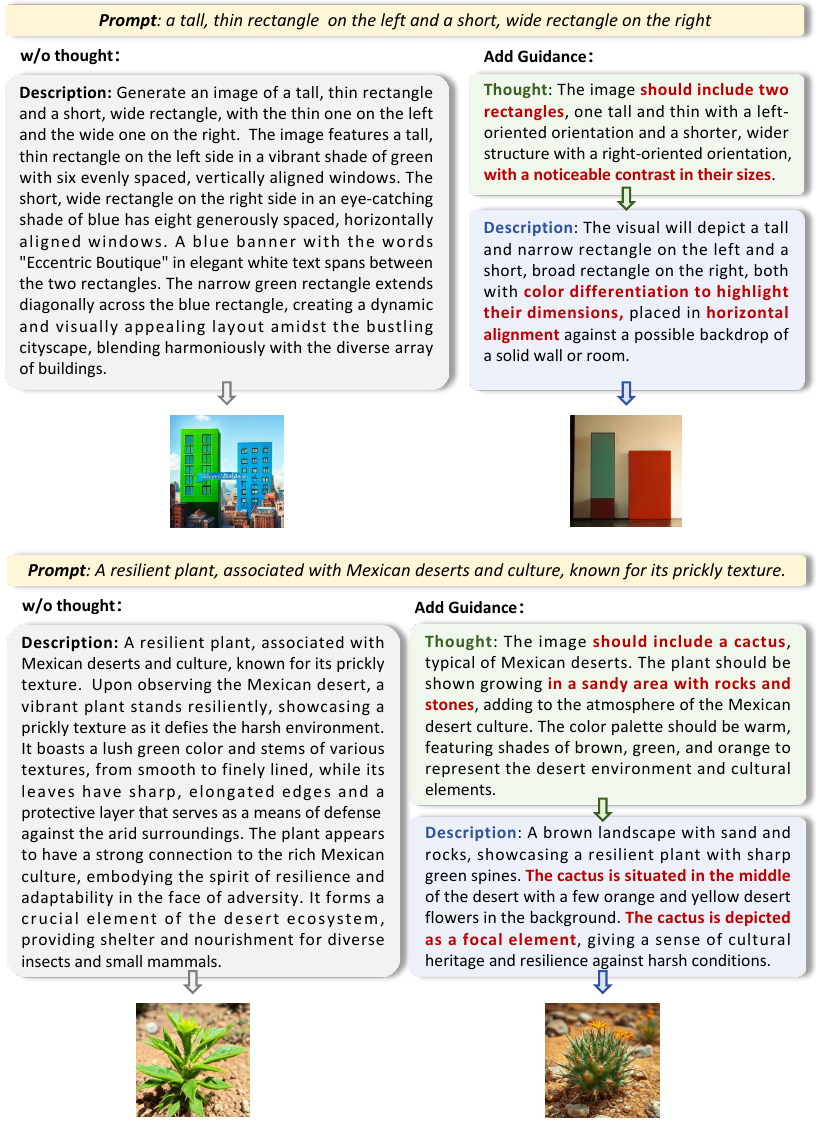}
  \caption{  Comparisons of images with and without the thought process.
 %
  }
  \label{fig:case6}
\end{figure*}

\begin{figure*}[t]
  \centering
    \includegraphics[width=0.85\linewidth]{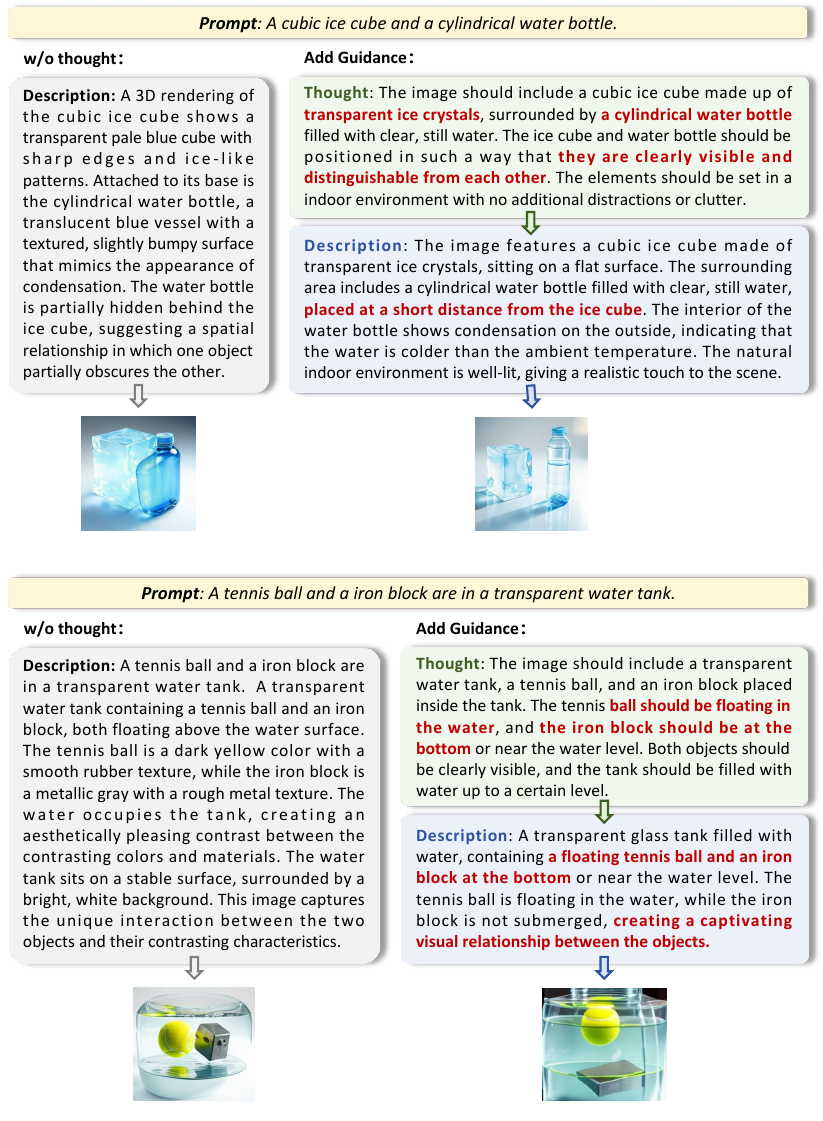}
  \caption{  Comparisons of images with and without the thought process.
 %
  }
  \label{fig:case7}
\end{figure*}

\begin{figure*}[t]
  \centering
    \includegraphics[width=0.85\linewidth]{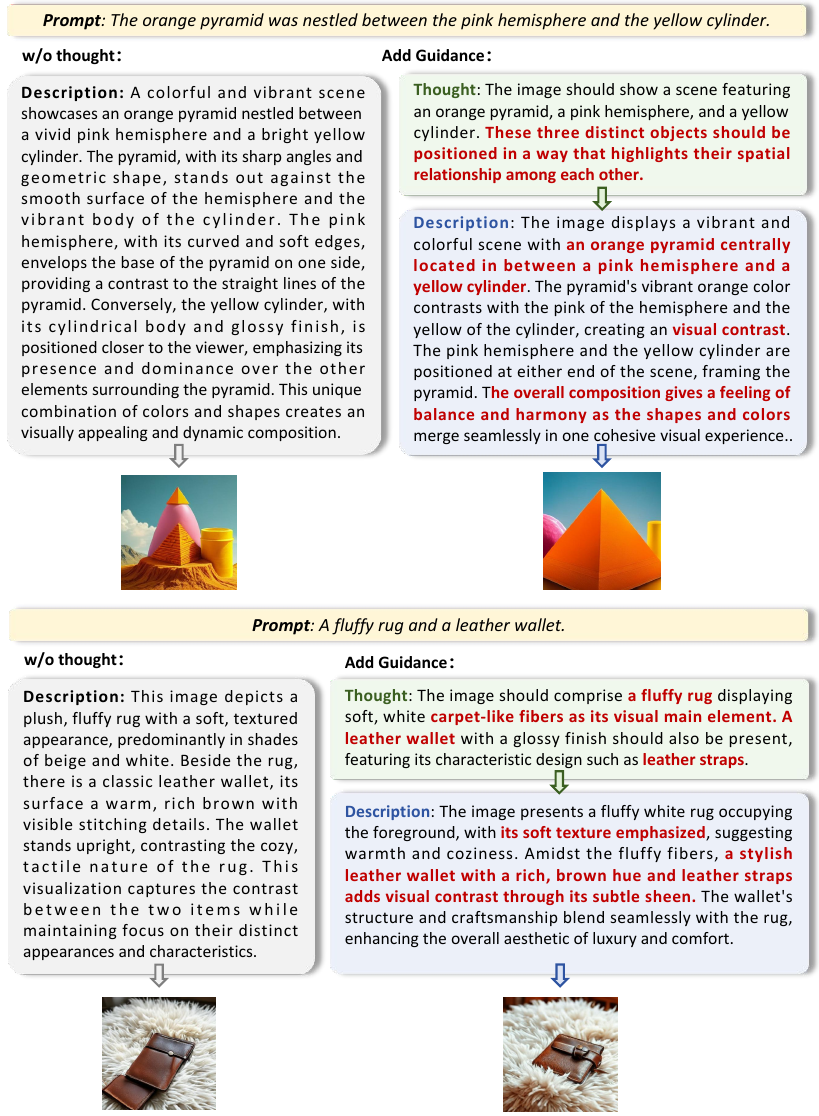}
  \caption{   Comparisons of images with and without the thought process.
 %
  }
  \label{fig:case8}
\end{figure*}

\end{document}